\def\year{2020}\relax
\documentclass[letterpaper]{article} % DO NOT CHANGE THIS
\usepackage{aaai20}  % DO NOT CHANGE THIS
\usepackage{times}  % DO NOT CHANGE THIS
\usepackage{helvet} % DO NOT CHANGE THIS
\usepackage{courier}  % DO NOT CHANGE THIS
\usepackage[hyphens]{url}  % DO NOT CHANGE THIS
\usepackage{graphicx} % DO NOT CHANGE THIS
\usepackage{graphicx}  % DO NOT CHANGE THIS
\usepackage{amsmath}
\usepackage{amssymb}
\usepackage{amsmath}  
\title{AdaFilter: Adaptive Filter Fine-tuning for Deep Transfer Learning }
\author{Yunhui Guo $^{\dagger}$ \quad  Yandong Li $^{\ddagger}$ 
\quad   Liqiang Wang$^{\ddagger}$ \quad Tajana Rosing $^{\dagger}$ \\
{University of California, San Diego, CA} $^{\dagger}$  
\quad University of Central Florida,
Orlando, FL $^{\ddagger}$  \\
yug185@eng.ucsd.edu, lyndon.leeseu@outlook.com,
lwang@cs.ucf.edu,
tajana@ucsd.edu}
 
\begin{document}

\maketitle

\begin{abstract}
There is an increasing number of pre-trained deep neural network models. However, it is still unclear how to effectively use these models for a new task. \textit{Transfer learning}, which aims to transfer knowledge from source tasks to a target task, is an effective solution to this problem. Fine-tuning is a popular transfer learning technique for deep neural networks where a few rounds of training are applied to the parameters of a pre-trained model to adapt them to a new task. Despite its popularity, in this paper we show that fine-tuning suffers from several drawbacks. We propose an adaptive fine-tuning approach, called AdaFilter, which selects only a part of the convolutional filters in the pre-trained model to optimize on a \textit{per-example} basis. We use a recurrent gated network to selectively fine-tune convolutional filters based on the activations of the previous layer. We experiment with 7 public image classification datasets and the results show that AdaFilter can reduce the average classification error of the standard fine-tuning by 2.54\%.

\end{abstract}

\section{Introduction}
% \textit{Inductive transfer learning} \cite{pan2010survey,torrey2010transfer}
\textit{Inductive transfer learning} \cite{pan2010survey}
is an important research topic in traditional machine learning that attempts to develop algorithms to transfer knowledge from \textit{source tasks} to enhance learning in a related \textit{target task}. While transfer learning for traditional machine learning algorithms has been extensively studied \cite{pan2010survey}, how to effectively conduct transfer learning using deep neural networks is still not fully exploited. The widely adopted approach, called \textit{fine-tuning}, is to continue the training of a pre-trained model on a given target task. Fine-tuning assumes the source and target tasks are related. Thus, we would expect the pre-trained parameters from the source task to be close to the optimal parameters for the target target.

In the standard fine-tuning, we either optimize all the pre-trained parameters or freeze certain layers (often the initial layers) of the pre-trained model and optimize the rest of the layers towards the target task. This widely adopted approach has two potential issues. First, the implicit assumption behind fine-tuning is that all the images in the target dataset should follow the same fine-tuning policy (i.e., all the images should fine-tune all the parameters of the pre-trained model), which may not be true in general. Recent work
% \cite{rosenbaum2017routing,shazeer2017outrageously} point out that per-example
\cite{shazeer2017outrageously} points out that activating parts of the network on a \textit{per-example} basis can better capture the diversity of the dataset to provide better classification performance. In transfer learning, some classes in the target dataset may have overlap with the source task, so directly reusing the pre-trained convolutional filters without fine-tuning can benefit the learning process due to the \textit{knowledge transfer}. The second issue is that deep neural networks often have millions of parameters, so when the pre-trained model is being applied on a relatively small target dataset, there is a danger of \textit{overfitting} since the model is \textit{overparameterized} for the target dataset. Solving either of the issues is of great practical importance due to the wide use of fine-tuning for a variety of applications. In a recent work \cite{guo2019spottune}, the authors proposed an adaptive fine-tuning method to generate fine-tuning policy for a new task on a per-example basis. However, the generated fine-tuning policy can be only used to decide which layers to be fine-tuned. 

In this paper, we propose a deep transfer learning model, called AdaFilter, which automatically selects reusable \textit{filters} from the pre-trained model on a \textit{per-example} basis. This is achieved by using a recurrent neural network (RNN) gate \cite{graves2013speech}, which is conditioned on the activations of the previous layer, to layerwisely decide which filter should be reused and which filter should be further fine-tuned for each example in the target dataset. In AdaFilter, different examples in the target dataset can fine-tune or reuse different convolutional filters in the pre-trained model. AdaFilter mitigates the overfitting issue by reducing the number of trainable parameters for each example in the target dataset via reusing pre-trained filters. We experiment on 7 publicly available image classification datasets. The results show that the proposed AdaFilter outperforms the standard fine-tuning on all the datasets.

The contributions of this paper can be summarized as follows, 
\begin{itemize}
    \item We propose AdaFilter, a deep transfer learning algorithm which aims to improve the performance of the widely used fine-tuning method. We propose \textit{filter selection}, \textit{layer-wise recurrent gated network} and \textit{gated batch normalization} techniques to allow different images in the target dataset to fine-tune different convolutional filters in the pre-trained model. 
    
    \item We experiment with 7 publicly available datasets and the results show that the proposed method can reduce the average classification error by 2.54\% compared with the standard fine-tuning.
    
    \item We also show that AdaFilter can consistently perform better than the standard fine-tuning during the training process due to more efficient \textit{knowledge transfer}.
\end{itemize}

\section{Related Work}

Transfer learning addresses the problem of how to transfer the knowledge from source tasks to a target task
% \cite{pan2010survey,csurkasurvey}. Transfer learning is closely related to
\cite{pan2010survey}. It is closely related to
life-long learning \cite{parisi2019continual}, multi-domain learning \cite{guo2019depthwise} and multi-task learning \cite{ruder2017overview}. Transfer learning has achieved great success in a variety of areas, such as computer  vision \cite{raina2007self}, recommender systems \cite{pan2010survey} and natural language processing \cite{min2017question}. Traditional transfer learning approaches include subspace alignment \cite{gong2012geodesic}, instance weighting
% \cite{fernando2013unsupervised,gong2012geodesic}, instance weighting
\cite{dudik2006correcting} and model adaptation \cite{duan2009domain}. 

% Recently, a slew of works \cite{ganin2016domain,co-regularized,ge2017borrowing}
Recently, a slew of works~\cite{co-regularized,ge2017borrowing,li2018hierarchical,li-etal-2019-transferable} are focusing on transfer learning with deep neural networks. The most widely used approach, called fine-tuning, is to slightly adjust the parameters of a pre-trained model to allow \textit{knowledge transfer} from the source task to the target task. Commonly used fine-tuning strategies include fine-tuning all the pre-trained parameters and fine-tuning the last few layers \cite{long2015learninglastfew}. In \cite{yosinski2014transferable}, the authors conduct extensive experiments to investigate the transferability of the features learned by deep neural networks. They showed that transferring features is better than random features even if the target task is distant from the source task. Kornblith et al. \cite{kornblith2018better} studied the problem that whether better ImageNet models transfer better. Howard et al. \cite{howard2018universal} extended the fine-tuning method to natural language processing (NLP) tasks and achieved the state-of-the-art results on six NLP tasks. In \cite{ge2017borrowing}, the authors proposed a selective joint fine-tuning scheme for improving the performance of deep learning tasks with insufficient training data. In order to get the parameters of the fine-tuned model close to the original pre-trained model, \cite{li2018explicit} explicitly added regularization terms to the loss function. In \cite{guo2019spottune}, the authors proposed a method to decide which layers to be fine-tuned for a new task. The proposed AdaFilter is more fine-grained as the fine-tuning policy is filter-level rather than layer-level. 

%The proposed AdaFilter is complementary to previous works. For example, the works that are based on fine-tuning can utilize AdaFilter to further improve the results. In AdaFilter, we allow different examples in the target dataset to fine-tune different convolutional filters in the pre-trained model. The adaptive fine-tuning scheme takes the similarity between the source task and the target task into consideration. The examples in the target dataset which are similar to the source task can reuse more pre-trained filters to achieve better knowledge transfer. 

\section{AdaFilter}
\begin{figure*}[t]
\centering
     \includegraphics[width=0.9\textwidth]{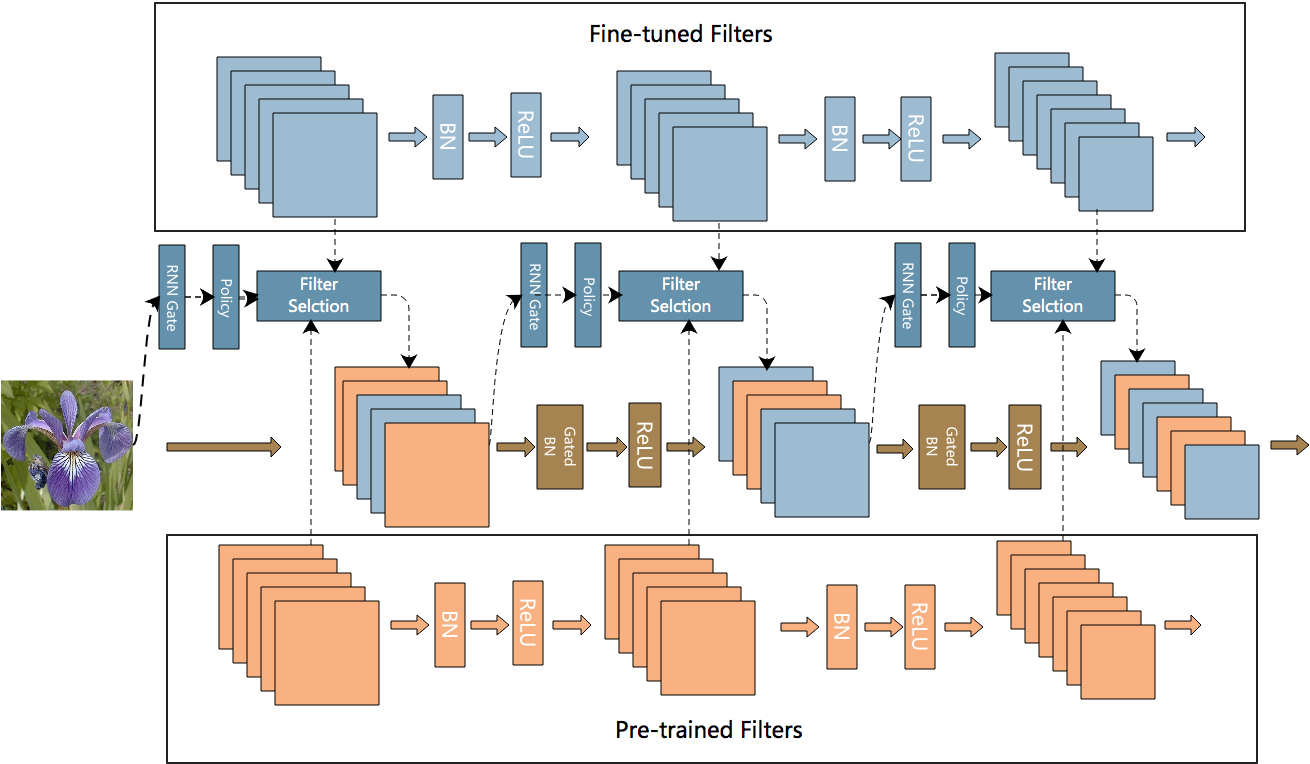}
\caption{The overview of AdaFilter for deep transfer learning. Layerwise recurrent gated network (the middle row) decides how to select the convolutional filters from the fine-tuned layer (the first row) and the pre-trained layer (the last row) conditioned on activations of the previous layer. Note that the RNN gate is shared across all the layers.}
\label{fig: convs}
\end{figure*}

In this work, we propose a deep transfer learning method which finds the convolutional filters in a pre-trained model that are reusable for each example in the target dataset. Figure~\ref{fig: convs} shows the overview of the proposed approach. We first use a \textit{filter selection} method to achieve per-example fine-tuning scheme. We therefore leverage a \textit{recurrent gated network} which is conditioned on the activations of the previous layer to layerwisely decide the fine-tuning policy for each example. Finally, we propose \textit{gated batch normalization} to consider the different statistics of the output channels produced by the pre-trained layer and the fine-tuned layer. 

To this end, we first introduce the filter selection method in Sec. 3.1. Then we introduce our recurrent gated network in
Sec. 3.2. Finally we present the proposed gated batch normalization method in Sec. 3.3. In Sec. 3.4, we discuss how the proposed method can achieve better performance by mitigating the issues of the standard fine-tuning.

% \begin{figure*}[t]
%   \centering
%   \includegraphics[width = 1\textwidth]{curve.pdf}

%   \caption{Transferabilities of BPDA (left) and  (right). Each entry shows the attack success rate of attacking the column-wise defense by the adversarial examples that are originally generated for the row-wise DNN.}
%   \label{fig:curve}
%   \vspace{-10pt}
% \end{figure*}

\subsection{Filter Selection}
In convolutional neural network (CNN), convolutional filters are used for detecting the presence of specific features or patterns in the original images. The filters in the initial layers of CNN are used for detecting low level features such as edges or textures while the filters at the end of the network are used for detecting shapes or objects \cite{zeiler2014visualizing}. When convolutional neural networks are used for transfer learning, the pre-trained filters can be reused to detect similar patterns on the images in the target dataset. For those images in the target dataset that are similar to the images in the source dataset, the pre-trained filters should not be fine-tuned to prevent from being destructed. 

 The proposed \textit{filter selection} method allows different images in the target dataset to fine-tune different pre-trained convolutional filters. Consider the $i$-th layer in a convolutional neural network with input feature map $x_i \in \mathbb{R}^{n_i \times w_i \times h_i}$, where $n_i$ the number of input channels, $w_i$ is the width of the feature map and $h_i$ is the height of the feature map. Given $x_i$, the convolutional filters in the layer $i$ produce an output $x_{i+1} \in \mathbb{R}^{n_{i+1} \times  w_{i+1} \times h_{i+1}}$. This is achieved by applying $n_{i+1}$ convolutional filter $\mathcal{F} \in \mathbb{R}^{n_{i} \times k \times k}$ on the input feature map. Each filter  $\mathcal{F} \in \mathbb{R}^{n_{i} \times k \times k}$ is applied on $x_i$ to generate one channel of the output. All the $n_{i+1}$ filters in the $i$-th convolutional layer can be stacked together as a $4D$ tensor.

We denote the $4D$ convolutional filters in the $i$-th layer as $F_i$. Given $x_i \in \mathbb{R}^{n_i \times w_i \times h_i}$, $F_i(x_i)$ is the output 
$x_{i+1} \in \mathbb{R}^{n_{i+1} \times  w_{i+1} \times h_{i+1}}$. To allow different images to fine-tune different filters, we initialize a new $4D$ convolutional filter $S_i$ from $F_i$ and \textit{freeze} $S_i$ during training. We use a binary vector $G_i(x_i)$ $\in \{0 ,1\}^{n_{i+1}}$, called the fine-tuning policy, which is conditioned on the input feature map $x_i$ to decide which filters should be reused and which filters should be fine-tuned. With $G_i(x_i)$, the output of the layer $i$ can be calculated as,

\begin{equation}
    x_{i+1} = G_i(x_i) \circ F_i(x_i) + (1-G_i(x_i)) \circ S_i(x_i)
\end{equation}

 where $\circ$ is a Hadamard product. Each element of $G_i(x_i)$ $\in \{0 ,1\}^{n_{i+1}}$ is multiplied with the corresponding channel of $F_i(x_i)  \in \mathbb{R}^{n_{i+1} \times  w_{i+1} \times h_{i+1}}$ and $S_i(x_i)  \in \mathbb{R}^{n_{i+1} \times  w_{i+1} \times h_{i+1}}$. Essentially, the fine-tuning policy $G_i(x_i)$ selects each channel of $x_{i+1}$ either from the output produced by the pre-trained layer $S_i$ (if the corresponding element is 0) or the fine-tuned layer $F_i$ (if the corresponding element is 1). Since $G_i(x_i)$ is conditioned on $x_i$, different examples in the target dataset can fine-tune different pre-trained convolutional filters in each layer.

\begin{figure}[!h]
\centering
     \includegraphics[width=0.4\textwidth]{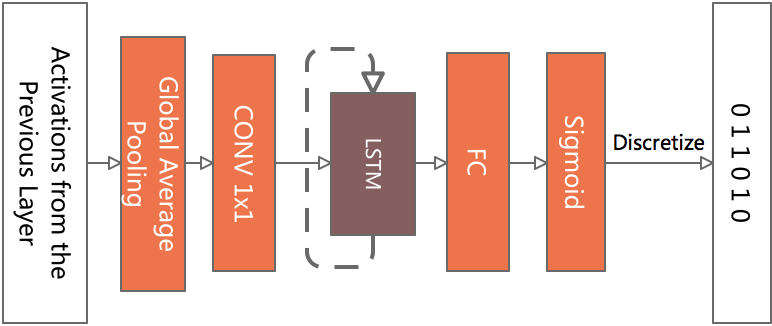}
\caption{The proposed recurrent gated network.}
\label{fig: rnn}

\end{figure}

\subsection{Layerwise Recurrent Gated Network}
There are many possible choices to generate the fine-tuning policy $G_i(x_i)$. We adopt a recurrent gated network to both consider the dependencies between different layers and the model size. Figure~\ref{fig: rnn} illustrates the proposed recurrent gated network which takes activations from the previous layer as input and discretizes the output of sigmoid function as the fine-tuning policy.

Recurrent neural network (RNN) \cite{graves2013speech} is a powerful tool for modelling sequential data. The hidden states of the recurrent neural network can remember the correlations between different timestamps. In order to apply the RNN gate, we need to map the input feature map $x_i$ into a low-dimensional space. We first apply a global average pooling on the input feature map $x_i$ and then use a $1 \times 1$ convolution which translates the $3D$ input feature map into a one-dimensional embedding vector. The embedding vector is used as the input of the RNN gate to generate the layer-dependent fine-tune policy $G_i(x_i)$. We translate the output of the RNN gate using a linear layer followed by a sigmoid function. To obtain the binary fine-tuned policy $G_i(x_i)$, we use a hard threshold function to discretize the output of the sigmoid function. 

The discreteness of $G_i(x_i)$ makes it hard to optimize the network using gradient-based algorithm. To mitigate this problem, we use the \textit{straight-through estimator} which is a widely used technique for training binarized neural network in the field of neural network quantization \cite{guo2018survey}. In the \textit{straight-through estimator}, during the forward pass we discretize the sigmoid function using a threshold and during backward we compute the gradients with respect to the input of the sigmoid function,
\begin{equation}
\begin{split}
& \textbf{Forward:}\quad x_b  =\begin{cases}
    1, & \textnormal{sigmoid}(x) \ge 0.5, \\
    0, & \text{otherwise}
  \end{cases} \\
& \textbf{Backward:}\quad \frac{\partial E}{\partial x}  =  \frac{\partial E}{\partial x_b}
% & \textbf{Backward:}\quad \frac{\partial x_b}{\partial x}  	
\end{split}
\end{equation}

\noindent where $E$ is the loss function. The adoption of the \textit{straight-through estimator} allows us to back-propagate through the discrete output and directly use gradient-based algorithms to end-to-end optimize the network. In the experimental section, we use Long Short-Term Memory (LSTM) \cite{hochreiter1997long} which has shown to be useful for different sequential tasks. We also compare the proposed recurrent gated network with a CNN-based gated network. The experimental results show that we can achieve higher classification accuracy by explicitly modelling the cross-layer correlations.

\subsection{Gated Batch Normalization}
Batch normalization (BN) \cite{ioffe2015batch} layer is designed to alleviate the issue of internal covariate shifting of training deep neural networks. In BN layer, we first standardize each feature in a mini-batch, then scale and shift the standardized feature. Let $X_{p,n}$ denote a mini-batch of data, where $p$ is the batch size and $n$ is the feature dimension. BN layer normalizes a feature dimension $x_j$ as below,

\begin{equation}
    \hat{x}_j = \frac{x_j - E[X_{.j}]}{ \sqrt{Var[X_{.j}]}}
\end{equation}
\begin{equation}
    y_j = \gamma_j \hat{x}_j + \beta_j
\end{equation}

The scale parameter $\gamma_j$ and shift parameter $\beta_j$ are trained jointly with the network parameters. In convolutional neural networks, we use BN layer after each convolutional layer and apply batch normalization over each channel \cite{ioffe2015batch}. Each channel has its own scale and shift parameters. 

In the standard BN layer, we compute the mean and variance of a particular channel across all the examples in the mini-batch. In AdaFilter, some examples in the mini-batch use the channel produced by the pre-trained filter while the others use the channel produced by the fine-tuned filter. The statistics of the channels produced by the pre-trained filters and the fine-tuned filters are different due to \textit{domain shift}. To consider this fact, we maintain two BN layers, called Gated Batch normalization, which normalize the channels produced by the pre-trained filters and the fine-tuned filters separately. To achieve this, we apply the fine-tuning policy learned by the RNN gate on the output of the BN layers to select the normalized channels,

\begin{equation}
    x_{i+1} = G_i(x_i) \circ BN_1(x_{i+1}) + (1-G_i(x_i))\circ BN_2(x_{i+1})
\label{Eq: Bns}
\end{equation}
where $BN_1$ and $BN_2$ denote the BN layer for the pre-trained filters and the fine-tuned filters separately and $\circ$ is the Hadamard product. In this way, we can deal with the case when the target dataset and the source dataset have very different domain distribution by adjusting the corresponding shift and scale parameters. 

\subsection{Discussion}
The design of the proposed AdaFilter mitigates two issues brought by the standard fine-tuning. Since the fine-tuning policy is conditioned on the activations of the previous layer, different images can fine-tune different pre-trained filters. The images in the target dataset which are similar to the source dataset can reuse more filters from the pre-trained model to allow better \textit{knowledge transfer}. On the other hand, 
while the number of parameters compared with standard fine-tuning increase by a factor of 2.2x with AdaFilter, the trainable parameters for a particular image are much fewer than the standard fine-tuning due to the reuse of the pre-trained filters. This alleviates the issue of \textit{overfitting} which is critical if the target dataset is much smaller than the source dataset. 

All the proposed modules are differentiable which allows us to use gradient-based algorithm to end-to-end optimize the network. During test time, the effective number of filters for a particular test example is equal to a standard fine-tuned model, thus AdaFilter has similar test time compared with the standard fine-tuning.

\section{Experimental Settings}

\subsection{Datasets}
We compare the proposed AdaFilter method with other fine-tuning and regularization methods on 7 public image classification datasets coming from different domains: Stanford dogs \cite{KhoslaYaoJayadevaprakashFeiFei_FGVC2011}, Aircraft \cite{maji2013fine}, MIT Indoors \cite{quattoni2009recognizing}, UCF-101 \cite{bilen2016dynamic}, Omniglot \cite{lake2015human}, Caltech 256 - 30 and Caltech 256 - 60 \cite{griffin2007caltech}. For Caltech 256 - $x$ ($x$ = 30 or 60), there are $x$ training examples for each class. The statistics of the datasets are listed in Table 1. Performance is measured by classification accuracy on the evaluation set.

\begin{table}[t]
\center
\small

\begin{tabular}{ |c|c|c|c| } 
 \hline
 \textbf{Dataset} &  \textbf{Training} & \textbf{Evaluation} & \textbf{Classes} \\ 
\hline
Stanford Dogs & 12000 & 8580 & 120\\
\hline
UCF-101 & 	7629 & 	1908 & 101 \\
\hline
Aircraft & 	3334 &3333& 100  \\

\hline
Caltech 256 - 30 & 7680 & 5120& 256 \\
\hline
Caltech 256 - 60 & 15360 & 5120 &256 \\

\hline
MIT Indoors & 5360 & 1340 & 67\\
\hline
Omniglot &	19476  &	6492 & 	1623\\
\hline
\end{tabular}

\caption{Datasets used to evaluate AdaFilter against other fine-tuning baselines.}
\label{table: datasets}
\end{table}

\def\arraystretch{1.0}%  1 is the default, change whatever you need
\begin{table*}[!tb]
\small
	\begin{center}
		\begin{tabular}{c c c c c c c c c} 
			\hline
		    Method  & Stanford-Dogs & UCF-101 & Aircraft  & Caltech256-30 & Caltech256-60 & MIT Indoors & Omniglot  \\
			\hline
			
%     	    Feature Extractor  &  & 34.73\% & 19.17\%  &  &   &  & 36.41\%  \\
% 			\hline 

    	    Standard Fine-tuning  & 77.47\% & 73.10\% & 52.59\%  & 78.09\% &  82.25\% & 76.42\% & 87.06\% \\
			\hline 

%     	    Fine-tuning last-1  &  & 72.33\%  &  47.61\%   & &   &  & 85.82\%  \\
% 			\hline 

%     	    Fine-tuning last-2  &  & 75.25\% & 48.21\%  & &   &  & 86.82\% \\
% 			\hline 
	
%     	    Fine-tuning last-3  &  &   76.07\% & 48.75\%   & &   &  & 86.95\% \\
% 			\hline 

			Fine-tuning half  & 79.61\%  &   76.43\% & 53.61\%  & 78.86\% & 82.55\%   &76.94\%  &  87.29\%\\
			\hline 
	
             Random Policy & 81.84\% &  75.15\% & 54.15\%  &79.90\%  &83.35\% &   76.71\% & 85.78\%  \\
			\hline 
	
    	    L2-SP &  79.69\% & 74.33\%  & \textbf{56.52\% } & 79.33\%  & 82.89\%  &  76.41\%& 86.92\% \\
			\hline 
			
    	    \textbf{AdaFilter}  & \textbf{82.44\%} & \textbf{76.99\%} & 55.41\%  & \textbf{80.62\%}& \textbf{84.31\%} &  \textbf{77.53\%}& \textbf{87.46\%}  \\
			\hline 
		\end{tabular}
	\end{center}
		\caption{The results of AdaFilter and all the baselines.}
	\label{table:results}
\end{table*}

\subsection{Baselines}
We consider the following fine-tuning variants and regularization techniques for fine-tuning in the experiments,

\begin{itemize}
    \item Standard Fine-tuning: this is the standard fine-tuning method which fine-tunes all the parameters of the pre-trained model.
    \item Fine-tuning half: only fine-tune second half of the layers of the pre-trained model and freeze the first half of layers.
    \item Random Policy: use AdaFilter with a random fine-tuning policy. This shows the effectiveness of fine-tuning policy learned by the recurrent gated network.
    \item L2-SP \cite{li2018explicit}: this is a recently proposed regularization method for fine-tuning which explicitly adds regularization terms in the loss function to encourage the fine-tuned model to be similar to the pre-trained model.
\end{itemize}

\subsection{Pretrained Model }
To compare AdaFilter with each baseline. We use ResNet-50 which is pre-trained on ImageNet. The ResNet-50 starts with a convolutional layer followed by 16 blocks with residual connection. Each block contains three convolutional layers and are distributed into 4 macro blocks (i.e, [3, 4, 6, 3]) with downsampling layers in between. The ResNet-50 ends with an average pooling layer followed by a fully connected layer. For a fair comparison with each baseline, we use the pre-trained model from Pytorch which has a classification accuracy of 75.15\% on ImageNet.

\subsection{Implementation Details}
Our implementation is based on
Pytorch. All methods are trained on 2 NVIDIA Titan Xp GPUs. We use SGD with momentum as the optimizer. The initial learning rate is 0.01 for the classification network and the initial learning rate for the recurrent gated network is 0.1. The momentum rate is 0.9 for both classification network and recurrent gated network. The batch size is 64. We train the network with a total of 110 epochs. The learning rate decays three times at the 30th, 60th and 90th epoch respectively.

\begin{figure*}[t]
  \centering
  \includegraphics[width = 0.9\textwidth]{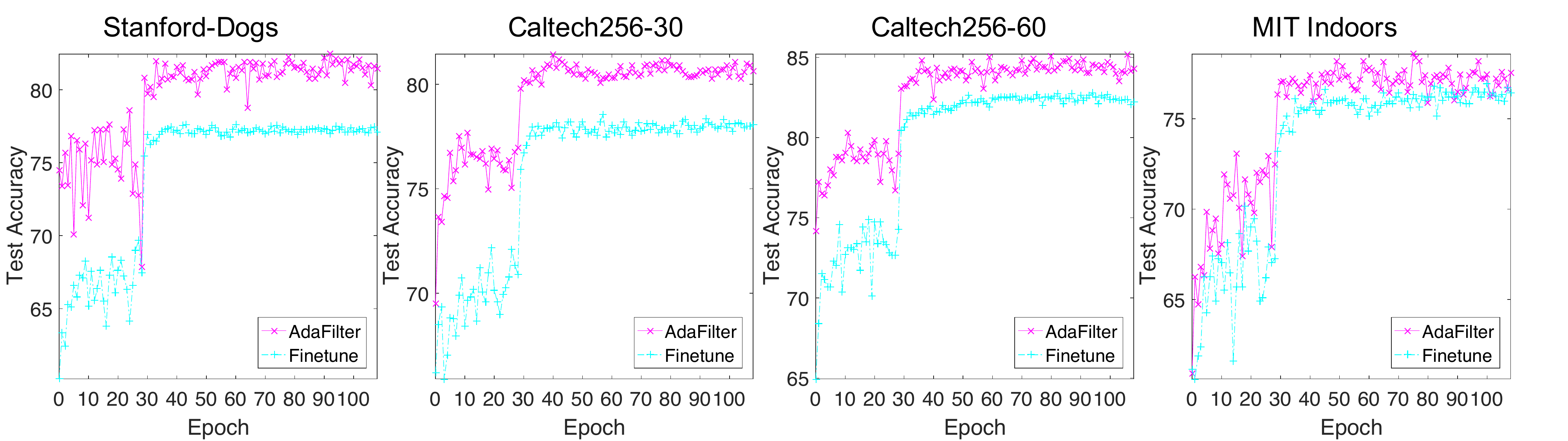}

  \caption{The test accuracy curve of AdaFilter and the standard fine-tuning on Stanford-Dogs, Caltech256-30, Caltech256-60 and MIT Indoors.}
  \label{fig:curve}
  \vspace{-10pt}
\end{figure*}

\section{Results and Analysis}

\subsection{Quantitative Results}

\subsubsection{AdaFilter vs Baselines }
We show the results of AdaFilter and all the baselines in Table 2. AdaFilter achieves the best results on 6 out of 7 datasets. It outperforms the standard fine-tuning on all the datasets. Compared with the standard fine-tuning, AdaFilter can reduce the classification error by up to 5\%. This validates our claim that by exploiting the idea of per-example filter fine-tuning, we can greatly boost the performance of the standard fine-tuning method by mitigating its drawbacks. While fine-tuning half of the layers generally performs better than the standard fine-tuning, it still performs worse than AdaFilter since it still applies the same fine-tuning policy for all the images which ignores the similarity between the target task and the source task.

Compared with Random policy and L2-SP, AdaFilter obtains higher accuracy by learning optimal fine-tuning policy for each image in the target dataset via the recurrent gated network. The results reveal that by carefully choosing different fine-tuning  for different images in the target dataset, we can achieve better transfer learning results. 
With AdaFilter, we can automatically specialize the fine-tuning policy for each test example which cannot be done manually due to the huge search space.

\begin{figure*}[!h]
  \centering
  \includegraphics[width = 0.48\textwidth]{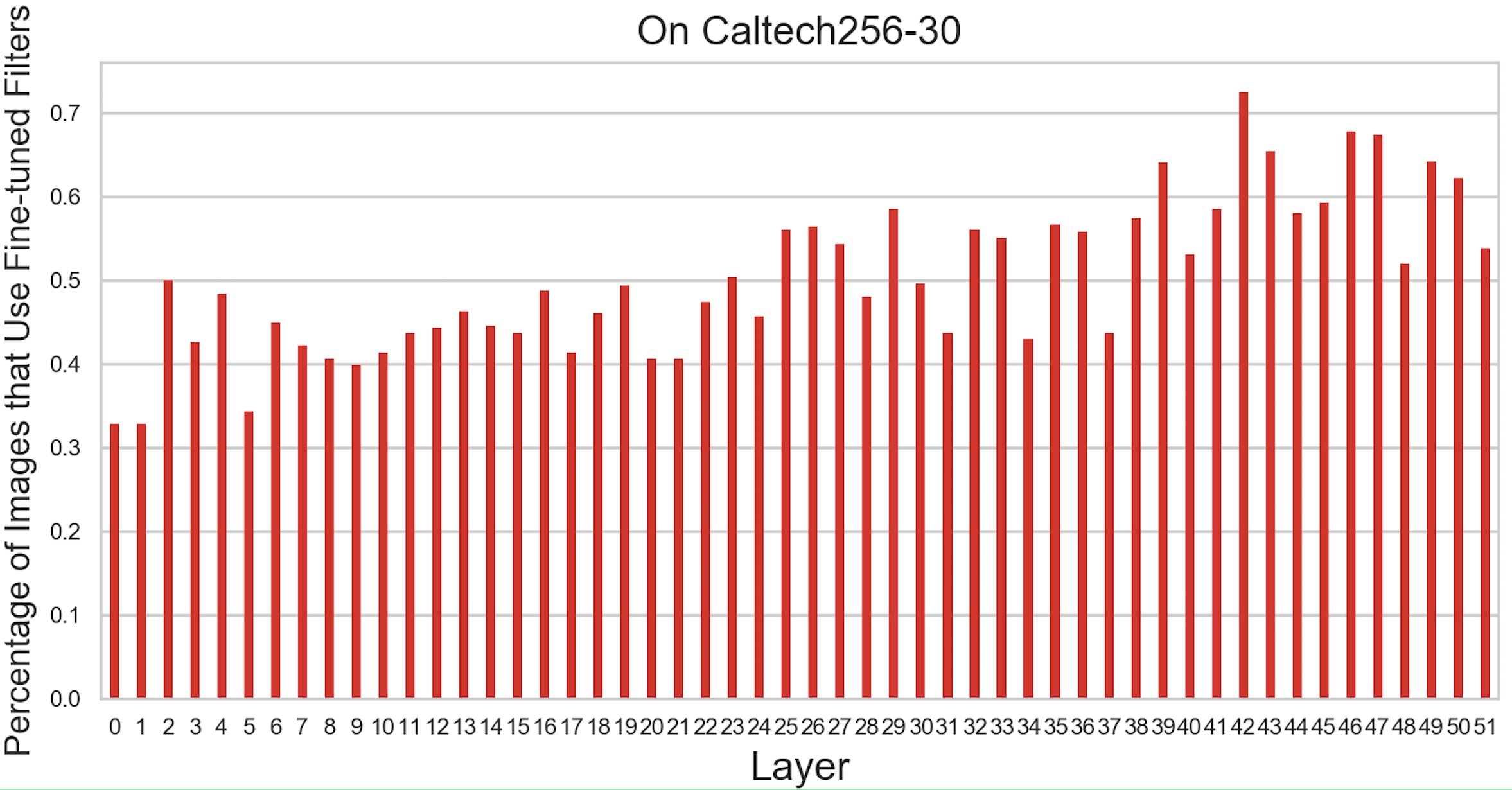}
  \includegraphics[width = 0.48\textwidth]{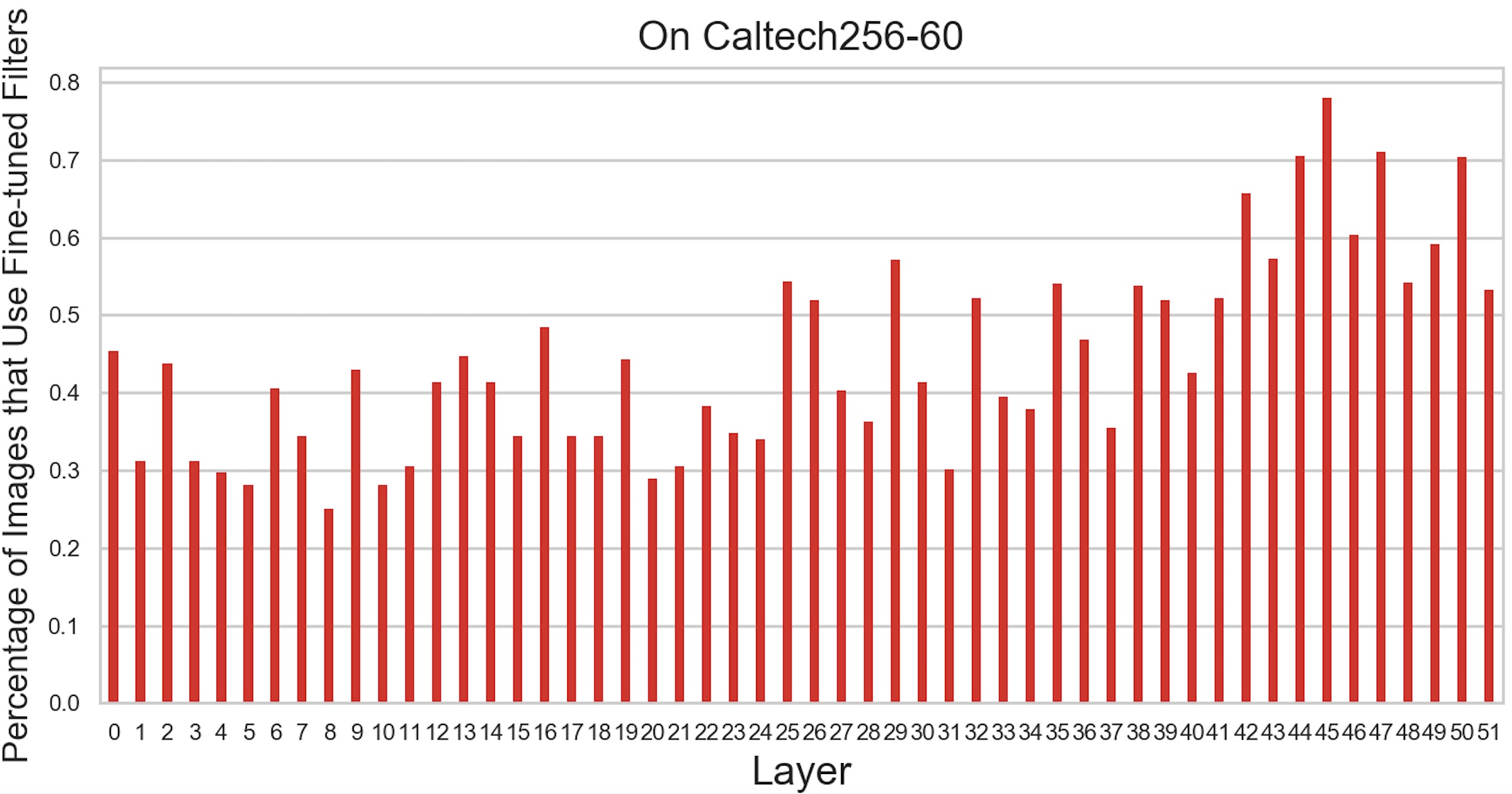}
  \caption{The visualization of fine-tuning policies on Caltech256-30 and Caltech256-60. }
  \label{fig:policies}
  \vspace{-10pt}
\end{figure*}

\subsubsection{Test accuracy curve} We show the test accuracy curve on four benchmark datasets in Figure \ref{fig:curve}. We can clearly see that the proposed AdaFilter consistently achieves higher accuracy than the standard fine-tune method across all the datasets. For example, after training for one epoch, AdaFilter reaches a test accuracy of 71.96\% on the Stanford Dogs dataset while the standard fine-tuning method only achieves 54.69\%. Similar behavior is also observed on other datasets. The fact that AdaFilter can reach the same accuracy level as standard fine-tuning with much fewer epochs is of great practical importance since it can reduce the training time on new tasks.

% \begin{figure*}[!htb]
% \minipage{0.25\textwidth}
%   \includegraphics[width=\linewidth]{dogs.eps}
% \endminipage\hfill
% \minipage{0.25\textwidth}
%   \includegraphics[width=\linewidth]{cal30.eps}
% \endminipage\hfill
% \minipage{0.25\textwidth}%
%   \includegraphics[width=\linewidth]{cal60.eps}
% \endminipage\hfill
% \minipage{0.25\textwidth}%
%   \includegraphics[width=\linewidth]{mitindoor.eps}
% \endminipage
%   \caption{A really Awesome Image}
% \end{figure*}

\subsection{Qualitative Results}

\subsubsection{Visualization of Policies}

In this section, we show the fine-tuning policies learned by the recurrent gated network on Caltech256-30 and Caltech256-60 in Figure~\ref{fig:policies}. The x-axis denotes the layers in the ResNet-50. The y-axis denotes the percentage of images in the evaluation set that use the fine-tuned filters in the corresponding layer. As we can see, there is a strong tendency for images to use the pre-trained filters in the initial layers while fine-tuning more filters at the higher layers of the network. This is intuitive since the filters in the initial layers can be reused on the target dataset to extract visual patterns (e.g., edges and corners). The higher layers are mostly task-specific which need to be adjusted further for the target task \cite{lee2008sparse}. We also note that the policy distribution is varied across different datasets, this suggests that for different datasets it is preferable to design different fine-tuning strategies.

\subsection{Ablation Study}

\subsubsection{Gated BN vs Standard BN}

In this section, an ablation study is performed to demonstrate the effectiveness of the proposed \textit{gated batch normalization}. We compare gated batch normalization (Gated BN) against the standard batch normalization (Standard BN). In Gated BN, we normalize the channels produced by the pre-trained filters and fine-tuned filters separately as in Equation \ref{Eq: Bns}. In standard batch normalization, we use one batch normalization layer to normalize each channel across a mini-batch,

\begin{equation}
    x_{i+1} = BN(x_{i+1})
\end{equation}

Table 3 shows the results of the Gated BN and the standard BN on all the datasets. Clearly, Gated BN can achieve higher accuracy by normalizing the channels produced by the pre-trained filters and fine-tuned filters separately. This suggests that although we can reuse the pre-trained filters on the target dataset, it is still important to consider the difference of the domain distributions between the target task and the source task.

\begin{table}[t]
\center
\small
\resizebox{\columnwidth}{!}{
\begin{tabular}{ |c|c|c|c|c|c|c|c| } 
 \hline
 \textbf{Dataset} & Stanford-Dogs &  Aircraft & Omniglot & UCF-101 &MIT Indoors &Caltech-30 &Caltech-60\\ 
\hline
Gated BN &  82.44\% & 55.41\% &87.46\% & 76.99\% &77.53\% &80.06\% & 84.31\%\\
\hline
Standard BN & 82.02\% &54.33\% & 87.27\% & 76.02\% & 77.01\% & 79.84\% & 83.84\% \\

\hline
\end{tabular}
}
\caption{Comparison of Gated BN and the standard BN}
\label{table: bns}
\end{table}

\subsubsection{Recurrent Gated Network vs CNN-based Gated Network}

In this section, we perform an ablation study to show the effectiveness of the proposed recurrent gated network. We compare the recurrent gated network against a CNN-based policy network. The CNN-based policy network is based on ResNet-18 which receives images as input and predicts the fine-tuning policy for all the filters at once. In the CNN-based model, the input image is directly used as the input for the CNN. The output of the CNN is a list of fully connected layers (one for each output feature map in the original backbone network) followed by sigmoid activation function. We show the results of the recurrent gated network and CNN-based policy network in Table~\ref{table: rnn}. Recurrent gated network performs better than the CNN-based policy network on most of the datasets by explicitly considering the dependency between layers. More importantly, predicting the policy layerwisely and reusing the hidden states of the recurrent gated network can greatly reduce the number of parameters. The lightweight design of the recurrent gated network is also faster to train than the CNN-based alternative.

\begin{table}[t]
\center
\small
\resizebox{\columnwidth}{!}{
\begin{tabular}{ |c|c|c|c|c|c|c|c|c| } 
 \hline
 \textbf{Dataset} & Stanford-Dogs &  Aircraft & Omniglot & UCF-101&MIT Indoors &Caltech-30 &Caltech-60 \\ 
\hline
RNN-based &  82.44\% & 55.41\% &87.46\% & 76.99\% &77.53\% &80.06\% & 84.31\% \\
\hline
CNN-based &  83.05\% & 54.63\% & 87.04\% & 76.33\% &77.46\% & 80.25\% &83.41\% \\

\hline
\end{tabular}
}
\caption{Comparison of recurrent gated network and a CNN-based policy network. The ``RNN-based'' means the recurrent gated network.}
\label{table: rnn}
\end{table}

\section{Conclusion}
In this paper, we propose a deep transfer learning method, called AdaFilter, which adaptively fine-tunes the convolutional filters in a pre-trained model. With the proposed \textit{filter selection}, \textit{recurrent gated network} and \textit{gated batch normalization} techniques, AdaFilter allows different images in the target dataset to fine-tune different pre-trained filters to enable better knowledge transfer. We validate our methods on seven publicly available datasets and show that AdaFilter outperforms the standard fine-tuning on all the datasets. The proposed method can also be extended to lifelong learning \cite{yoon2017lifelong} by modelling the tasks sequentially.

\section{Acknowledgment}
This work is supported in part by CRISP, one of six centers in JUMP, an SRC program sponsored by DARPA. This work is also supported by NSF CHASE-CI \#1730158, NSF \#1704309 and Cyber Florida Collaborative Seed Award.

% \bibliography{ref}
\bibliographystyle{aaai}

\end{document}